\documentclass[journal,10pt]{IEEEtran}
\IEEEoverridecommandlockouts
\pdfoutput=1
\usepackage{cite}
\usepackage{amsmath,amssymb,amsfonts}
\usepackage{algorithm}
\usepackage{algpseudocode}
\usepackage{subcaption}
\usepackage{graphicx}
\usepackage{textcomp}
\usepackage{array}
\usepackage{colortbl}
\usepackage[dvipsnames]{xcolor}
\usepackage{placeins} 
\usepackage{tcolorbox}
\usepackage{multirow}
\usepackage{booktabs}
\usepackage{fancyhdr}
\makeatletter
\newcommand*{\rom}[1]{\expandafter\@slowromancap\romannumeral #1@}
\makeatother
\def\BibTeX{{\rm B\kern-.05em{\sc i\kern-.025em b}\kern-.08em
    T\kern-.1667em\lower.7ex\hbox{E}\kern-.125emX}}

\usepackage{listings}
\definecolor{codegreen}{rgb}{0,0.6,0}
\definecolor{codegray}{rgb}{0.5,0.5,0.5}
\definecolor{codepurple}{rgb}{0.58,0,0.82}
\definecolor{backcolour}{rgb}{0.92,0.92,0.92}

\begin{document}


\title{SANGAM: \textbf{S}ystemVerilog \textbf{A}ssertio\textbf{n} \textbf{G}eneration vi\textbf{a} \textbf{M}onte Carlo Tree Self-Refine}

\author{\IEEEauthorblockN{Adarsh Gupta*, Bhabesh Mali*, Chandan Karfa}\\
\IEEEauthorblockA{Indian Institute of Technology Guwahati, India\\
\ \{adarsh.gupta, m.bhabesh, ckarfa\}@iitg.ac.in}}

\maketitle

\begin{abstract}

Recent advancements in the field of reasoning using Large Language Models (LLMs) have created new possibilities for more complex and automatic Hardware Assertion Generation techniques. This paper introduces SANGAM, a SystemVerilog Assertion Generation framework using LLM-guided Monte Carlo Tree Search for the automatic generation of SVAs from industry-level specifications. The proposed framework utilizes a three-stage approach: Stage 1 consists of multi-modal Specification Processing using Signal Mapper, SPEC Analyzer, and Waveform Analyzer LLM Agents. Stage 2 consists of using the Monte Carlo Tree Self-Refine (MCTSr) algorithm for automatic reasoning about SVAs for each signal, and finally, Stage 3 combines the MCTSr-generated reasoning traces to generate SVA assertions for each signal. The results demonstrated that our framework, SANGAM, can generate a robust set of SVAs, performing better in the evaluation process in comparison to the recent methods. 
\end{abstract}


\section{Introduction}
\label{sec:Introduction}
In recent years, the use of LLMs has surged in Electronic Design Automation (EDA), because of their capabilities in natural language understanding, reasoning, and automated code generation. Hardware Assertion Generation using various methods incorporating LLMs, \textcolor{black}{has attracted the major attention of researchers} \cite{llmsva1, mali2024chiraag}. SystemVerilog Assertions (SVAs) play a critical role in Assertion-Based Verification (ABV) for ensuring the correctness and security of hardware designs. Manually writing SVAs is time-consuming for verification engineers and often becomes more complex for larger designs, and can also lead to various errors due to language ambiguity. 

In this paper, we propose SANGAM, a framework that uses LLM-guided Monte Carlo Tree Search (MCTS) to automate SVA generation from multi-modal specifications. Using LLMs for MCTS requires various approximations in reward calculations and node expansion. SANGAM iteratively explores the solution space by continuously refining assertions based on critic and formal feedback. This iterative exploration allows SANGAM to explore various possible reasoning paths, helping capture a wide range of SVAs.

\def\thefootnote{*}\footnotetext{These authors contributed equally to this work}\def\thefootnote{\arabic{footnote}}

We developed a three-stage strategy to produce the assertions. The first stage involves processing the multi-modal specification data, consisting of text, waveforms, and images. It involves three custom LLMs: Spec Analyzer, Signal Mapper, and Waveform Analyzer. Each LLM produces significant and essential information about every signal and waveform using the design specification. Then stage two and three use this information to generate the SVAs. Stage two incorporates a modified version of the MCTSr Algorithm \cite{MCTSr} to produce various reasoning paths for the SVAs in a world model given by the LLM. Stage three further de-duplicates the assertions in all the reasoning paths and combines them into a cohesive set of assertions. \textcolor{black}{The code and information are available at online\footnote{https://github.com/CoolSunflower/SANGAM}}. Specifically, the key contributions of our work are:

\begin{itemize}
    \item We proposed a novel framework, SANGAM, to generate SVAs from multi-modal specifications consisting of design documents, architecture, and waveforms.
    \item SANGAM consists of three main steps: Specification Processing, Assertion Reasoning, and Assertion Generation, focusing on signal-wise reasoning and SVA generation. It also uses a novel \texttt{Waveform Analyzer LLM} during Specification Processing to extract signal interdependence from the specification waveforms.
    \item SANGAM framework uses a novel MCTS-based algorithm for effective exploration of the assertion space for each signal. This effective exploration allows it to discover an extensive set of assertions for each signal.
    \item SANGAM generates twice the assertions as state-of-the-art methods, such as AssertLLM \cite{assertllm} and ChIRAAG \cite{mali2024chiraag}. \textcolor{black}{ The coverage analysis demonstrates that these assertions are of high quality.}
\end{itemize}

The rest of the paper is organized as follows: Section \rom{2} covers the background and related works. Section \rom{3} covers the proposed methodology. Section \rom{4} covers the experimental results and discussion. Section \rom{5} covers the conclusion and future works.

\section{Background and Related Works}
\subsection{LLM for EDA}


The growth of the LLM industry is estimated to increase exponentially from USD 6.4 billion in 2024 to USD 36.1 billion by 2030 \cite{LLMReport}. LLMs are becoming an integral part of the EDA flow. As in \cite{xiong2024hlspilot}, the authors have introduced the first LLM-enabled high-level synthesis (HLS) framework, HLSPilot. The framework fully automates the HLS code generation, utilizing sequential C/C++ code as input to the LLM. \textcolor{black}{While \cite{reddy2024lhs} have used LLMs to reduce manual efforts while designing ML hardware accelerators.} In \cite{ho2024large}, the authors introduce a novel and efficient LLM model designed especially for standard cell layout design optimization for generating high-quality cluster constraints to enhance cell layout performance, power, and area (PPA). \textcolor{black}{While \cite{ blocklove2025automatically, vijayaraghavan2024vhdl} shows the capability of LLMs in generating HDL Code.} LayoutCopilot \cite{liu2025layoutcopilot} is the first interactive framework for LLM-powered analog layout design. 

\subsection{LLM for Assertion Generation}
Manually writing SVAs is a time-consuming process. Recently researchers have focused on generating assertions using LLMs because of their exceptional ability to generate quality results. In \cite{mali2024chiraag}, the authors have proposed a framework named ChIRAAG that automates the process of generating assertions using OpenAI's GPT4 model. They provided natural language specification as input to the LLM, further evaluating the generated assertions using testbench, and re-prompted the model to address any error that occurred during evaluation. 

In \cite{fang2024assertllm}, the authors proposed AssertLLM, a framework that utilizes three different LLMs for generating SVAs. Firstly, natural language specifications are analyzed using \texttt{Spec Analyzer LLM} to extract information for each signal. \textcolor{black}{The \texttt{Signal Mapper LLM} then maps each signal to its Verilog variable name using the provided Verilog code.} Finally, this information is given to an \texttt{SVA Generator LLM} to generate the SVAs. \textcolor{black}{However, since the assertions are generated in a single iteration, there is limited scope for refinement in case of errors. Furthermore, most assertions are not aligned with the design’s core functionality.} The authors of \cite{10458667} have generated hardware security assertions using LLM and also contributed to providing a comprehensive benchmark suite consisting of real-world hardware designs and corresponding golden assertions. They took comments from the assertion's code file and processed them to feed as input to the LLM. Their experiments showed that with sufficient details in the prompt, LLMs are capable of achieving good results in generating assertions.

\subsection{Monte Carlo Tree Search (MCTS)}
MCTS is a decision-making algorithm, widely used in games and complex decision processes. It operates by building a search tree and simulating outcomes to estimate the value of actions \cite{browne2012survey}. A single iteration of MCTS proceeds in the following four phases:
\begin{itemize}
    \item \textbf{Selection:} Starting from the root node, the algorithm goes through all the nodes and selects the most promising child node based on specific strategies, such as the Upper Confidence Threshold (UCT).
    \item \textbf{Expansion:} At this child node, a feasible child node is added that denotes a potential future move.
    \item \textbf{Simulation for Evaluation:} From this newly added node, the algorithm conducts a random simulation, by arbitrarily selecting moves until it reaches a terminal node. The reward provided at the terminal node is a measure of the goodness of this newly selected action.
    \item \textbf{Backpropagation:} Post-simulation, the results of the simulation are propagated back through all the parents to the root to inform future selection decisions.
\end{itemize}

MCTS is a general search algorithm, which operates in any search space to give optimal action paths. This search algorithm can be combined with the text-based model of the world provided by LLMs, to get a better reasoning model, one that optimally traverses the search space to find the answer. This has been explored in many works by various names, Monte Carlo Tree Self-refine (MCTSr) \cite{zhang2024accessing}, OmegaPRM \cite{luo2024improve}, SC-MCTS \cite{gao2024interpretable}.
Our work uses a modified version of the MCTSr algorithm, along with the open-source DeepSeek-R1 model \cite{DeepSeekR1} as the LLM world model, to get a better reasoning model than the base model. {\it The current literature does not explore this method of utilizing the MCT Self-Refine algorithm in a Multi-Round LLM environment for SVA Generation.}

\section{SANGAM}
\label{sec:SANGAM}

\begin{figure*}[htp]
    \centering
    \includegraphics[height=140pt,width=0.85\linewidth]{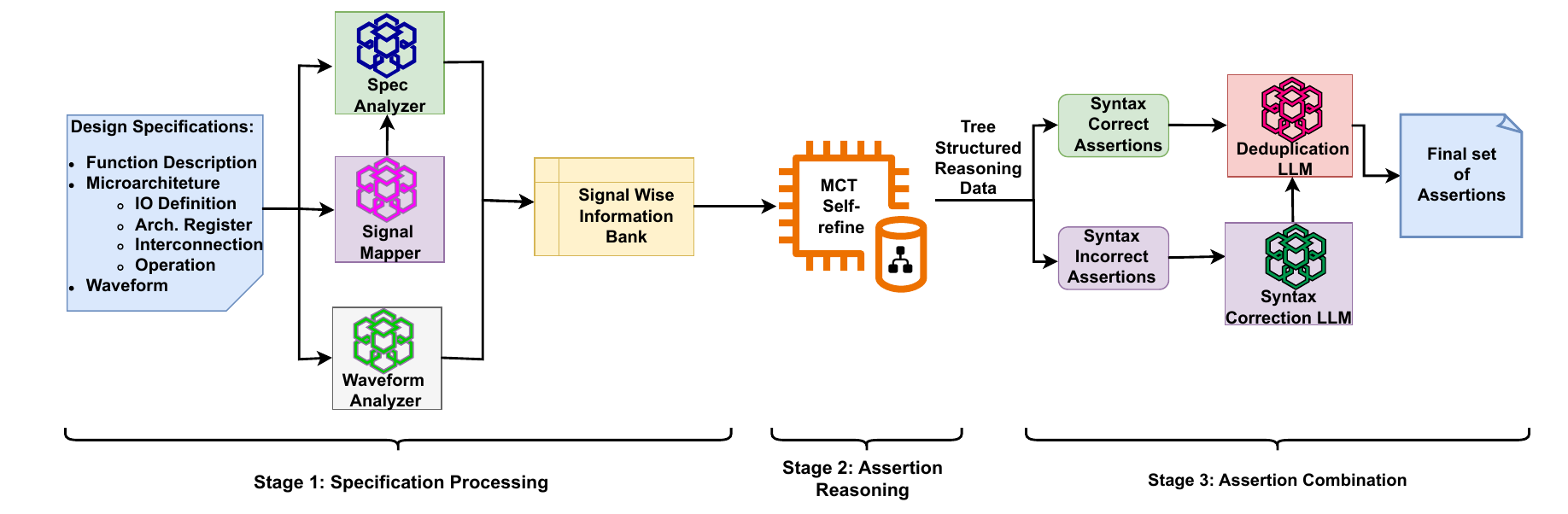}
    \caption{SANGAM: Proposed Architecture}
    \label{fig:enter-label}
\end{figure*}

Our proposed framework, SANGAM, is divided into three stages. These stages are depicted in Figure \ref{fig:enter-label}.

\subsection{Stage 1: Specification Processing}
The goal of this stage is to parse the multi-modal specification to produce a signal-wise information bank.
This stage involves the utilization of three different LLMs for specification processing. They are: \textbf{Signal Mapper}, \textbf{Spec Analyzer}, and \textbf{Waveform Analyzer}.  The signal-wise information bank contains the following information:

\begin{itemize}
    \item \textbf{Workflow Information:} Contains the signal mapping information, waveform analyses, and overall design summary from the architecture diagram.
    \item \textbf{Specification Analysis for each Signal:} Signal-wise information generated by the Spec Analyzer.
\end{itemize}


The role of all three  LLMs are discussed below:

\subsubsection{Signal Mapper}



The custom system instruction, shown in Figure \ref{fig: promptsignal}, instruct the \texttt{Signal Mapper LLM} to connect the specification file with the implementation or signal mapping details from the uploaded files and generate a brief description for each signal, along with the corresponding signal name used in the implementation, to facilitate easier signal mapping. The prompt also explicitly disallows hallucination and provides the required output format. 



\subsubsection{Spec Analyser}
The system prompt provided to the \texttt{Spec Analyzer LLM} instructs the model to act as a professional VLSI Specification Analyzer. The design specification file is provided as an input file to the LLM in PDF format. Following this, the system prompt instructs the LLM to structure its output for any particular signal extracted by the \texttt{Signal Mapper LLM} and to cover the signal name, and description containing the definition, functionality, interconnections, any additional points, and also information about related signals. Figure \ref{fig:prompt1} shows the prompts passed to the LLM.

\subsubsection{Waveform Analyzer}
The \texttt{Waveform Analyzer LLM} extracts signal interdependence information by analyzing the waveforms in the specification. This ensures that any properties missed by the Spec Analyzer are captured. The custom instruction, as shown in Figure \ref{fig: promptwave}, instructs the LLM to consider each of the provided waveform diagrams in context with the provided specification and generate a signal interdependence summary of the signals in the waveform.

The data generated by all the LLMs is combined to form the information bank, which is used in the next stages to generate signal-specific assertions by using signal-wise information and workflow information.

\subsection{Stage 2: Assertion Generation}
For each signal in the Information Bank, we run a modified version of MCTSr algorithm \cite{zhang2024accessing}, to construct a reasoning tree $R$ that captures the selection, expansion, and feedback processes in the form of a tree. In the reasoning tree, $R$, each node represents a set of assertions, and the children of that node represent possible improvements made to those assertions by incorporating feedback from a Critic and a formal verification tool.

\begin{figure*}[!ht]
    \centering
    \begin{subfigure}[b]{0.5\textwidth} 
        \centering
        \includegraphics[width=0.7\textwidth]{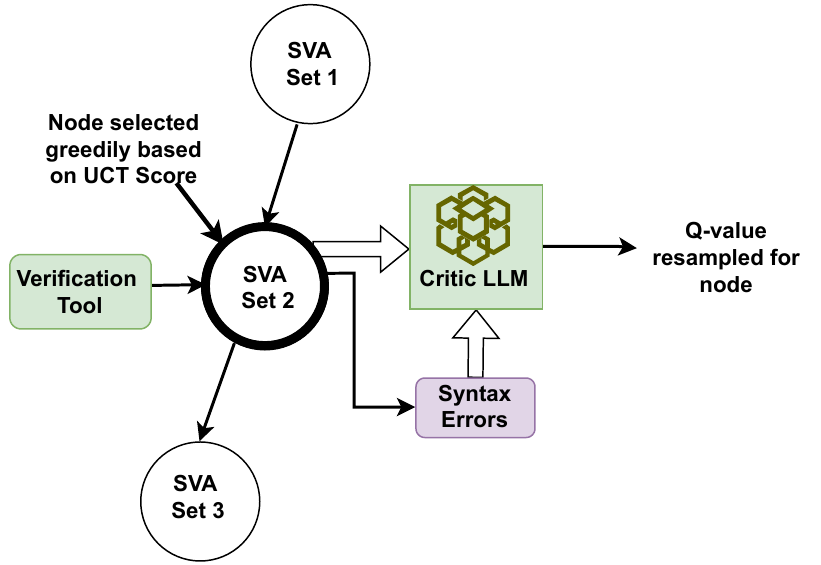}
        \caption{Phase 1: Selecting best nodes greedily using UCT score and resampling reward}
        \label{fig:subfig1}
    \end{subfigure}
    \hfill
    \begin{subfigure}[b]{0.45\textwidth} 
        \centering
        \includegraphics[width=0.85\textwidth]{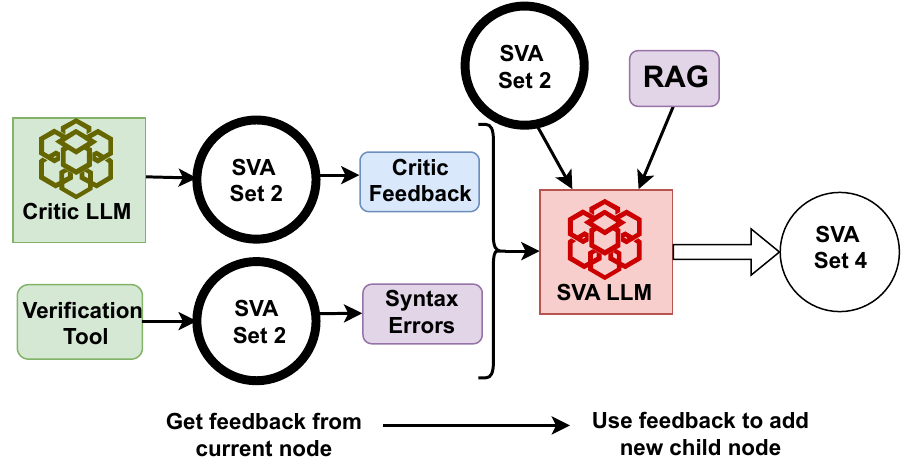}
        \caption{Phase 2: Node Expansion using Critic and Verification Tool Feedback}
        \label{fig:subfig2}
    \end{subfigure}

    \vspace{-1pt}  

    \begin{subfigure}[b]{0.4\textwidth} 
        \centering
        \includegraphics[width=\textwidth]{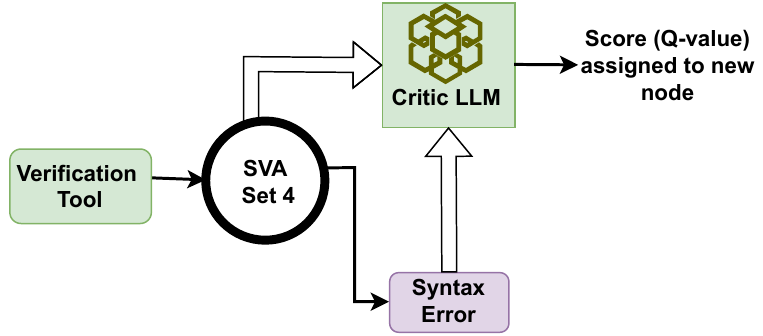}
        \caption{Phase 3: Node Evaluation}
        \label{fig:subfig3}
    \end{subfigure}
    \hspace{0.5cm} 
    \begin{subfigure}[b]{0.40\textwidth} 
        \centering
        \includegraphics[width=0.35\textwidth]{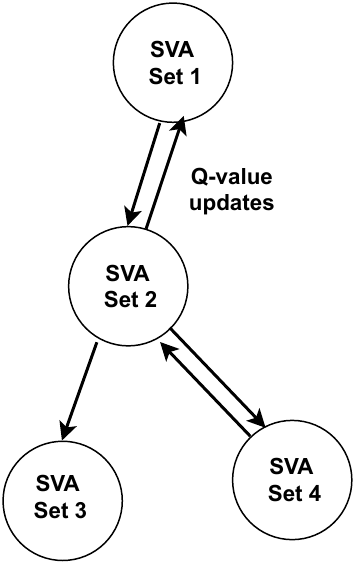}
        \caption{Phase 4: Backpropagation of score through tree}
        \label{fig:subfig4}
    \end{subfigure}

    \caption{Steps in each rollout of Assertion Generation}
    \label{fig:four-subfigs}
\end{figure*}

The typical MCTSr algorithm consists of four phases in each rollout: Node Selection, Node Expansion, Node Evaluation and Back-propagation.
The typical algorithm is modified by incorporating syntax feedback from a verification tool before a new set of assertions is produced in the Node Expansion step. This ensures that the framework can focus both on generating new assertions and ensuring the syntactic correctness of the old assertions.  This syntax feedback is also used for Reward Sampling during Node Evaluation and Node Selection.  The phases of the proposed MCTSr-based assertion generation algorithm are depicted in Figure \ref{fig:four-subfigs} and are discussed below.

\subsubsection{Node Selection}
Each node $a$ in the reasoning tree, $R$, is assigned a Q value ($Q(a)$) during evaluation in the previous rollouts which denotes the goodness of the assertions in that node as judged by a Critic. This Q value is used in combination with other factors like the number of times the node was visited earlier ($N(a)$), the number of times its parent node has been visited ($N(Father(a))$) to calculate the Upper Confidence Threshold ($UCT$) value for that node ($UCT_a$)

$$
UCT_a = Q(a) + c\sqrt{\frac{\text{ln}(N(\text{Father(}a\text{)})) + 1}{N(a) + \epsilon}},
$$

where $\epsilon$ is a small constant to avoid dividing by zero errors, $c$ is a constant used to balance exploration and exploitation and $N(\text{Father}(a))\text{ is equivalent to }$$N(Parent(a))$.

This UCT value assigned to any node signifies how good this node is for further exploration. We use greedy sampling to select the node with the highest UCT value for further exploration. The score of the selected node is sampled again and used to update its goodness value.

\subsubsection{Node Expansion (Self-Refine)}
Once a candidate node is selected, the next step is to expand the tree to add a child node. Expansion is a dual-phase process and is the key component for constructing the reasoning tree. It consists of generating feedback for the current set of assertions and then using this feedback to further refine the set of assertions. Two types of feedback are generated for the selected node which is used to create a new child node.


\textbf{Formal Verification Logs} are generated for each assertion in the current node. These logs contain the syntax verification information generated by Cadence JasperGold Formal Verification tool \cite{jg}. \textbf{Critic Feedback} is generated by a \texttt{Critic LLM} that is tasked with strictly evaluating and generating feedback for the assertions. The system prompt provided to this LLM is shown in Figure \ref{fig: critic_prompt}. An instance of the \texttt{Critic LLM} is created using the system prompt and is passed the Weak Answer of the node along with specification information to get the required feedback.

Following this, the query is augmented with additional context using Retrieval-Augmented Generation (RAG). A variety of documents \cite{Dasgupta2006}\cite{Mehta2013}\cite{Vijayaraghavan2005}
relevant to SVA generation are provided and the relevant snippets are extracted. \textcolor{black}{RAG is implemented \textcolor{black}{via a FAISS index on document chunks} to address LLM hallucination and provide better responses with respect to prompts.} Once the two feedbacks and additional context are obtained, they are used to generate a better set of assertions and subsequently a child node is added to the current node. The LLM responsible for generating these new assertions using feedback and old assertions is called the \texttt{SVA LLM}. The system prompt passed to the \texttt{SVA LLM} is shown in Figure \ref{fig:sva_prompt}. Using an instance of this \texttt{SVA LLM} we produce a new set of assertions and add a new node to the reasoning tree.

\subsubsection{Node Evaluation}
In Node Evaluation, the newly added node is assigned a goodness score. This is done by utilizing a feedback-aided self-reward method to estimate the reward between -100 to 100. The generated assertions are passed to the verification tool, Cadence JasperGold. The tool returns any syntax issue found in the generated assertions. This information is then passed on to the \texttt{Critic LLM} to generate a score between -100 and 100. The system prompt passed to the \texttt{Critic LLM} is depicted in Fig. \ref{fig: critic_prompt}. The feedback generated by the \texttt{Critic LLM} is discarded here.
It was found that in the absence of any prompt constraint, the output of the critic tends to be overly optimistic. To address this issue, we use three techniques:

\begin{itemize}
    \item \textit{Prompt Constraint:} The model must adhere to strict standards, while also focusing on the Completeness, Consistency, and Correctness of the generated SVAs.
    \item \textit{Full Score Suppression:} Scores above 95 are suppressed to curb overly optimistic model.
    \item \textit{Repeated Sampling:} The model score for any node is resampled if this node is selected again for expansion.
\end{itemize}

\subsubsection{Back-propogation}
After a new leaf node is added and assigned a score, we need to update the Q values of all the parents. This is done to reward the good paths of reasoning and increase its chance of selection in the next iteration. This updation is done with the following rule:

\begin{equation}
\label{eq:Q}
Q'(a) = \frac{1}{2} \left( Q(a) + \max_{i \in \text{children}(a)} Q(i) \right)    
\end{equation}
where $Q'(a)$ is the updated Q value of node $a$, $Q(i)$ is the Q value of any node $i$ in the tree.

This process is repeated $n\_rollouts$ times. The reasoning tree thus created, has a root node, with a weak assertion set and nodes added in each rollout, thus it consists of $n\_rollouts + 1$ nodes. After the reasoning tree is constructed for a particular signal, the last stage is to combine all the reasoning paths in $R$ to produce the final set of assertions for the signal.

\subsection{Stage 3: Assertion Combination}

The generated assertions in all the nodes of $R$ are collected and divided into two groups: Syntax Correct Assertions $A_1$ and Syntax Incorrect Assertions $A_2$ using the verification tool. The assertions in $A_2$ are provided to the \texttt{Correction LLM}, along with the system prompt shown in Figure \ref{fig:syntax_correction_llm}, to generate a syntactically correct set $A_2'$. The assertions in $A_1$ and $A_2'$ are combined to form $A_3$. 

The assertions in $A_3$ are passed into the \texttt{\texttt{De-duplication LLM}}, while the system prompt passed is given in Figure \ref{fig:deduplication}. The final set of assertions thus obtained, $A_{deduplicated}$, for each signal can be manually verified by the Verification Engineer against the specification to obtain the final correct set of assertions. This process of generating assertions is repeated for every signal produced by the \texttt{Signal Mapper LLM} during Specification Processing.

\section{Experimentation, Results and Discussion}
We have chosen two designs: Inter-Integrated Circuit (I2C), and RISC-V Timer (RV-Timer) to compare our results with the state-of-the-art methods, AssertLLM \cite{fang2024assertllm} and ChIRAAG \cite{mali2024chiraag}. We have used the DeepSeek-R1 model to create all the LLM agents for experimentation. \textcolor{black}{The cost analysis to generate the assertions for each signal is provided in Appendix B.}


\subsection{Implementation Details}

\textcolor{black}{After extracting the signal-wise information and signal interdependence information using the \texttt{SPEC Analyzer LLM} and \texttt{Waveform Analyser LLM}, all these details for each signal were combined to construct the Signal Wise Information Bank for each design. The signal interdependence information was not used for the RV-Timer design as the waveform information is missing in the specification. Then, for each signal of each particular design, we construct the reasoning tree, $R$, by iterative sampling and refinement.
We perform four rollouts ($n\_rollouts = 4$) to construct $R$. To initialize $R$, a weak answer node is added, whose content is generated using the \texttt{SVA LLM}, with additional prompt tuning to enforce the LLM to keep the generated answer short. This short weak answer node ensures that all the paths of reasoning starting at the root node will have a sufficient chance to be more creative rather than restricting their creative domain to the assertions in the root node. This weak answer node is evaluated and assigned a score between -100 and 100 using the node evaluation method as detailed earlier (Section III-B3). Following this, we proceed with each of the rollouts as shown in Fig. \ref{fig:four-subfigs}. However, in the first step of each rollout, the best node is selected greedily, for which exploration/exploitation parameter $c$ is set to 1.4. While, in the assertion combination step, the assertions in each node of $R$ are combined. Further, we have used the Cadence JasperGold Tool to divide the syntactically correct and incorrect assertions. The rest of the steps are discussed in Section III-C.}



\subsection{Results and Discussion}
Initially, our framework SANGAM generated a large number of assertions. We formally verified the generated assertions using the Cadence JasperGold Verification Tool. For each design, we have provided a detailed explanation below.

\subsubsection{I2C}

\begin{table}
\centering
\renewcommand{\arraystretch}{1.1}
\setlength{\tabcolsep}{6pt}
\begin{tabular}{>{\centering\arraybackslash}p{0.5cm}|>{\centering\arraybackslash}p{1cm}|>{\centering\arraybackslash}p{1.8cm}||>{\centering\arraybackslash}p{1.5cm}||>{\centering\arraybackslash}p{1.5cm}}
\toprule
\multicolumn{2}{c|}{\cellcolor{gray!10}\textbf{Signal Type}} & \cellcolor{gray!10}\textbf{Signal Name} & \cellcolor{gray!10}\textbf{SANGAM} & \cellcolor{gray!10}\textbf{AssertLLM} \\ 
\hline\hline
\multirow{17}{*}{\rotatebox[origin=c]{90}{\textbf{IO Port}}} & \textit{Clock} & wb\_clk\_i & 5 & 1 \\ \cline{2-5}
 & \multirow{2}{*}{\textit{Reset}} & wb\_rst\_i & 17 & 1 \\
 & & arst\_i & 13 & 1 \\ \cline{2-5}
 & \multirow{3}{*}{\textit{Control}} & wb\_stb\_i & 1 & 2 \\
 & & wb\_ack\_o & 9 & 1 \\
 & & wb\_inta\_o & 11 & 1 \\ \cline{2-5}
 & \multirow{11}{*}{\textit{Data}} & wb\_adr\_i & 1 & 1 \\
 & & wb\_dat\_i & 4 & 1 \\
 & & wb\_cyc\_i & 2 & 1 \\
 & & wb\_dat\_o & 3 & 1 \\
 & & wb\_we\_i & 4 & 1 \\
 & & scl\_pad\_i & 1 & 1 \\
 & & scl\_pad\_o & 6 & 1 \\
 & & sda\_pad\_i & 5 & 1 \\
 & & sda\_pad\_o & 4 & 1 \\
 & & scl\_pad\_oe & 5 & 1 \\
 & & sda\_pad\_oe & 18 & 1 \\ \hline
\multirow{6}{*}{\rotatebox[origin=c]{90}{\textbf{Register}}} & \multirow{2}{*}{\textit{Control}} & ctr & 7 & 10 \\
 & & sr & 12 & 14 \\ \cline{2-5}
 & \multirow{4}{*}{\textit{Data}} & prer & 5 & 2 \\
 & & txr & 10 & 2 \\
 & & rxr & 5 & 2 \\
 & & cr & 4 & 2 \\ \hline\hline
\multicolumn{3}{c||}{\cellcolor{gray!10}\textbf{Design Total (Complete Correct)}} & \multicolumn{1}{c||}{\cellcolor{gray!5}\textbf{152}} & \cellcolor{gray!5}\textbf{50} \\
\bottomrule
\end{tabular}

\caption{Evaluation of the generated SVAs for design ``I2C''. Signal wise comparison between SANGAM and AssertLLM \cite{fang2024assertllm}. Each entry denotes the number of correct assertions generated for that signal.}
\label{tbl:eval_i2c}
\end{table}

The generated assertions for each signal, after being deduplicated, were formally verified using the same RTL as provided in AssertLLM \cite{dataset}. We were able to identify a total of 152 syntactically and semantically correct assertions. This marks an increase of 204\% compared to the AssertLLM framework. The signal-wise analysis is given in Table \ref{tbl:eval_i2c}.

\subsubsection{RV-Timer}

\begin{table}
\centering
\renewcommand{\arraystretch}{1.1}
\setlength{\tabcolsep}{6pt}
\begin{tabular}{>{\centering\arraybackslash}p{0.5cm}|>{\centering\arraybackslash}p{1cm}|>{\centering\arraybackslash}p{1.8cm}||>{\centering\arraybackslash}p{1.5cm}||>{\centering\arraybackslash}p{1.5cm}}
\toprule
\multicolumn{2}{c|}{\cellcolor{gray!10}\textbf{Signal Type}} & \cellcolor{gray!10}\textbf{Signal Name} & \cellcolor{gray!10}\textbf{SANGAM} & \cellcolor{gray!10}\textbf{ChIRAAG} \\ 
\hline\hline
\multirow{7}{*}{\rotatebox[origin=c]{90}{\textbf{Input \& Registers}}} & \textit{Clock} & clk\_i & 2 & \\ \cline{2-4}
 & \multirow{1}{*}{\textit{Reset}} & rst\_i & 4 & \\ \cline{2-4}
 & \multirow{1}{*}{\textit{Active}} & active & 5 & \\ \cline{2-4}
 & \multirow{2}{*}{\textit{Timer}} & prescaler & 5 & \\
 & & step & 3 & $/$ \\ \cline{2-4}
 & \multirow{2}{*}{\textit{Storage}} & mtime & 4 & \\
 & & mtimecmp & 4 & \\ \cline{1-4}
\multirow{3}{*}{\rotatebox[origin=c]{90}{\textbf{Output}}} & \multirow{2}{*}{\textit{Output}} & tick & 8 & \\
 & & intr & 3 &  \\ \cline{2-4}
 & \multirow{1}{*}{\textit{Data}} & mtime\_d & 8 & \\ \hline\hline
\multicolumn{3}{c||}{\cellcolor{gray!10}\textbf{Design Total (Complete Correct)}} & \multicolumn{1}{c||}{\cellcolor{gray!5}\textbf{46}} & \cellcolor{gray!5}\textbf{11}\\
\bottomrule
\end{tabular}
\caption{Evaluation of SVAs for the "RV-Timer" design. Signal-wise comparison between SANGAM and ChIRAAG, with each entry showing the number of correct assertions per signal. (Signal-wise results are not available for ChIRAAG.)}

\label{tbl:eval_rv}
\end{table}

\begin{figure}[!t]
    \centering
    \includegraphics[width=0.7\linewidth]{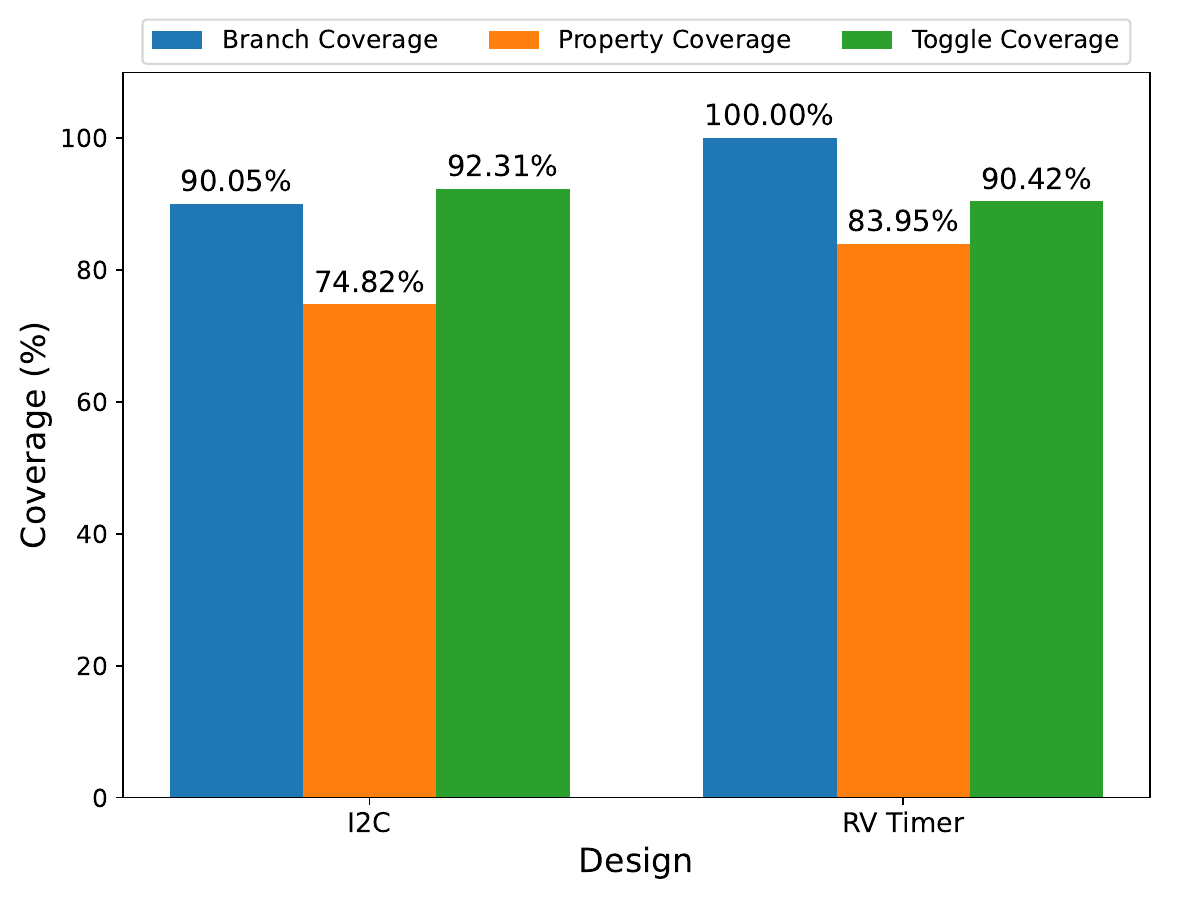}
    \caption{Coverage Analysis}
    \label{Cov}
\end{figure}

The generated assertions for each signal, after being deduplicated, were formally verified using the same RTL as used in ChIRAAG. We were able to identify a total 46 syntactically and semantically correct assertions. This marks an increase of 254\% compared to the ChIRAAG framework. The signal-wise analysis is given in Table \ref{tbl:eval_rv}. The results demonstrate that our proposed framework SANGAM provides extensive signal coverage by iteratively reasoning about the assertions for each signal. This significantly improves on the completeness aspect as compared to recent methods.
\textcolor{black}{We have also performed coverage analysis for both designs using the RTL and assertion files as inputs to the JasperGold Coverage Application. Fig. \ref{Cov} presents the resulting coverages for SANGAM. Both designs achieve above 90\% branch and toggle coverage and 74\% property coverage. Further, we performed the coverage analysis of the ChIRAAG-generated assertions for comparison. The property and toggle coverage achieved with ChIRAAG is 50\% and 92.07\%, respectively. While we found that the conditional statement is not covered as none of the assertions generated by ChIRAAG asserted the condition. As a result, the corresponding branch remained unproven during formal analysis.. These results demonstrate that SANGAM-generated assertions are more effective in capturing the design intent, taking a significant step toward completeness.}


A detailed comparison with AssertLLM is not feasible as the assertions generated by their method are not publicly available. However, for most of the signals, their framework was only able to generate assertions to check the width of the signal. In comparison, SANGAM was able to generate assertions that not only captured the width property but also more importantly the functional properties of the signals. However, we were able to give a detailed evaluation of the generated assertions of RV-Timer. The generated assertions capture various important characteristics of the design that are missed by ChIRAAG. For example, \textbf{Assertion 1} as shown below depicts the functional behavior that \texttt{intr} signal is asserted when the value in \texttt{mtime} register is greater than or equal to the value in \texttt{mtimecmp}. This behavior is a core functionality of the design and is not being captured by ChIRAAG \cite{mali2024chiraag}.

\begin{lstlisting}
(*@\textbf{Assertion 1:}@*) 
property mtime_intr_p;
  @(posedge clk_i) disable iff (!rst_ni)
  (mtime >= mtimecmp[0]) |-> intr[0];
endproperty
assert property (mtime_intr_p);
\end{lstlisting}


\section{Conclusion and Future Work}

This paper introduced SANGAM, a novel framework for automatic SVA generation using LLM-guided Monte Carlo Tree Search. By using a three-stage approach for each signal, SANGAM ensures extensive signal coverage, extracting all useful information from multi-modal specifications, and iteratively reasoning about assertions along multiple reasoning paths for improved correctness. Experimental results showcase a significant increase in the number of correct and functionally important assertions compared to existing methods. The generated assertions also demonstrated good performance in the coverage analysis. It will help the engineers to focus on validation rather than writing assertions from scratch. However, SANGAM is unable to guarantee inter-signal assertion irredundancy. Future work can focus on optimizing the computational efficiency and reducing the overhead from multiple LLM calls by using a domain-specific LLM instead of a general domain LLM. Further improvements can be made in fine-tuning the reward function used during node evaluation. 


\bibliographystyle{IEEEtran}
\vspace{-0.1 cm}
\bibliography{ref.bib}

\newpage
\appendices
\section{Custom Instructions}

\FloatBarrier 
\begin{figure}[hbtp!]
    \centering
    \begin{tcolorbox}[
        colback=lightgray!10, 
        colframe=black, 
        width=0.45\textwidth, 
        boxrule=0.8pt,  
        arc=5pt,  
        enlarge left by = 0.5mm, 
        fontupper=\small  
    ]
        {\fontsize{8pt}{8pt}\selectfont 
        \textbf{[System Prompt for Signal Mapper]}\\
        • Please act as a signal name mapping tool to link the specification file and the Verilog code.\\
        • Firstly, I’ll upload the the design specification file (in PDF format), and a Verilog file containing all the signal definitions (*.v format).\\
        • Only output a signal if it is present in both the specification and the Verilog file. Think step by step and verify your output before stopping.\\
        • The signal description should be short and to the point.\\ \\
        \textbf{[Instruction for Signal Mapper]}\\
        • Please analyze the specification file and use the code interpreter to analyze the Verilog file (both the signal declarations and comments). Then map every signal (including IO ports, wires, and registers) defined in Verilog with the description in the specification. Finally, please output each signal in the following format:

        [Signal name in Verilog]: Signal definition in Specification file
        }
    \end{tcolorbox}
    \caption{Custom Instructions for the \texttt{Signal Mapper LLM}}
    \label{fig: promptsignal}
\end{figure}

\begin{figure}[hbtp!]
    \centering
    \begin{tcolorbox}[
        colback=lightgray!10, 
        colframe=black, 
        width=0.45\textwidth, 
        boxrule=0.8pt,  
        arc=5pt,  
        enlarge left by = 0.5mm, 
        fontupper=\small  
    ]   {\fontsize{8pt}{8pt}\selectfont 
        \textbf{[System Prompt for Spec Analyzer]}\\
        • Please act as a professional VLSI specification analyzer.\\
        • Don't use any content outside the file for answering the questions. Think step by step.\\
        • When I ask for information on any signal, extract all the information, suitable for the SystemVerilog Assertions generation, and output all the information in the following format:
        
        [Signal Name]; [Description] $\rightarrow$ [Definition], [Functionality], [Interconnection], [Additional Information]; [Related Signals]. \\ \\
        \textbf{[Instruction for Spec Analyzer]}\\
        • \textit{(Upload the specification file)} Here is the design specification file, please analyze it carefully.\\
        • \textit{(For each signal)} Please extract all the information related to the SIGNAL NAME from the spec file.
        }
    \end{tcolorbox}
    \caption{Custom Instructions for the \texttt{Spec Analyzer LLM}}
    \label{fig:prompt1}
\end{figure}

\begin{figure}[htbp!]
    \centering
    \begin{tcolorbox}[
        colback=lightgray!10, 
        colframe=black, 
        width=0.45\textwidth, 
        boxrule=0.8pt,  
        arc=5pt,  
        enlarge left by = 0.5mm, 
        fontupper=\small
    ]

        {\fontsize{8pt}{8pt}\selectfont  
      
     • Please act as a critic to a professional VLSI verification engineer. You will be provided with a specification and workflow information. \\
     • Along with that, you will be provided with a signal name, its specification, and the SVAs generated by a professional VLSI verification engineer for that signal. \\
     • Analyze the SVAs strictly and criticize everything possible. You are not allowed to create new signal names apart from the specification. \\
     • Point out every possible flaw and give a score from -100 to 100. Also focus on Clock Cycle Misinterpretations, and nested if-else and long conditions. \\
     • Be very strict, ensure \textbf{CORRECTNESS}, \textbf{CONSISTENCY}, and \textbf{COMPLETENESS} of the  SVAs. Focus on these three things while grading the SVAs and providing feedback. \\
     • Let's think step by step.
    }  
    \end{tcolorbox}
    \caption{System Prompt for the \texttt{Critic LLM}}
    \label{fig: critic_prompt}
\end{figure}

\begin{figure}
    \centering
    \begin{tcolorbox}[
        colback=lightgray!10, 
        colframe=black, 
        width=0.45\textwidth, 
        boxrule=0.8pt,  
        arc=5pt,  
        enlarge left by = 0.5mm, 
        fontupper=\small  
    ]

        {\fontsize{8pt}{8pt}\selectfont  
        \textbf{[System Prompt for Waveform Analyzer]}\\
        • Please act as a professional waveform analyzer specialized in VLSI design verification. \\
        • Your primary task is to extract and summarize signal interdependence information from the waveform diagrams. \\  
        • For each waveform diagram, generate a structured signal interdependence summary in the following format:

        [Waveform Name]: Name as mentioned in the SPEC 
        
        [Signals]: List of all signals present in the waveform 
        
        [Interdependence Analysis]: 
        \begin{itemize}
            \item $[$Timing Relationship$]$: Define how signals interact over time (e.g., edge alignment, sequential dependencies, latency)
            \item $[$Causal Dependencies$]$: Identify signals that influence or trigger changes in others.
            \item $[$State Transitions$]$: Describe key transitions and conditions governing signal changes.
            \item $[$Protocol/Handshaking Mechanisms$]$: If applicable, explain interactions like request-acknowledge, arbitration, or synchronization.
            \item $[$Additional Observations$]$: Any other relevant information inferred from the waveform.\\
        \end{itemize}

        \textbf{[Instruction for Spec Analyzer]}\\
        • \textit{(Upload the specification file)} Here is the specification. Analyze it carefully.\\
        • \textit{(Upload each waveform as Image)} Extract information in specified format.
    }  
    \end{tcolorbox}
    \caption{Custom Instructions for \texttt{Waveform Analyzer LLM}}
    \label{fig: promptwave}
\end{figure}

\begin{figure}[htbp!]
    \centering
    \begin{tcolorbox}[
        colback=lightgray!10, 
        colframe=black, 
        width=0.45\textwidth, 
        boxrule=0.8pt,  
        arc=5pt,  
        enlarge left by = 0.5mm, 
        fontupper=\small
    ]

        {\fontsize{8pt}{8pt}\selectfont  
     • Please act as a professional VLSI verification engineer. You will be provided with a specification, workflow information, signal name and its description.\\
     • Please write all the corresponding SVAs based on the defined Verilog signals that benefit both the RTL design and verification processes.\\
     • Do not generate any new signals. Make sure that the generated SVAs have no syntax error and strictly follow the function of the given specification/description.\\
     •  The generated SVAs should include but not be limited to the following types: [width] [connectivity] [function]\\
     • Ensure that you do not create a new signal than the specification.\\
     • Sometimes you may also be provided with improvement feedback, take it into consideration and improve your SVAs while maintaining \textbf{CORRECTNESS}, \textbf{CONSISTENCY}, and \textbf{COMPLETENESS}. Let's think step by step.
     
    }  
    \end{tcolorbox}
    \caption{System Prompt for \texttt{SVA LLM}}
    \label{fig:sva_prompt}
\end{figure}

\begin{figure}[htbp!]
    \centering
    \begin{tcolorbox}[
        colback=lightgray!10, 
        colframe=black, 
        width=0.45\textwidth, 
        boxrule=0.8pt,  
        arc=5pt,  
        enlarge left by = 0.5mm, 
        fontupper=\small  
    ]   {\fontsize{8pt}{8pt}\selectfont 
        • Using the following documentation for reference return the assertions (provided later) with their syntax issue fixed: \{specification\_text\}. Here are the SystemVerilog assertions with syntax errors:\\
        • \textit{(added syntactically wrong assertions along with the syntax error found)}\\
        • Please correct the syntax of the assertions provided earlier and return only the corrected assertions for \{signal\_name\}. You are not allowed to create any new signals than the ones specified in the specification.
        }
    \end{tcolorbox}
    \caption{Custom Instructions for \texttt{Syntax Correction LLM}}
    \label{fig:syntax_correction_llm}
\end{figure}

\begin{figure}[htbp!]
    \centering
    \begin{tcolorbox}[
        colback=lightgray!10, 
        colframe=black, 
        width=0.45\textwidth, 
        boxrule=0.8pt,  
        arc=5pt,  
        enlarge left by = 0.5mm, 
        fontupper=\small  
    ]   {\fontsize{8pt}{8pt}\selectfont 
        • Using the following documentation for reference: \{specification\_text\}. Here are several SystemVerilog assertions: \{assertions\}\\
        • Extract all unique and valid assertions from this list for the \{signal\_name\}. Ensure that you maximise the number of retained assertions.\\
        • Do not attempt to modify the actual assertion in any way. Don't forget any assertion (width, connectivity, functionality).
        }    
    \end{tcolorbox}
    \caption{Custom Instructions for the \texttt{Deduplication LLM}}
    \label{fig:deduplication}
\end{figure}

\section{Cost Analysis}

Maximum number of API calls for each signal = 2 (For creating Initial Node of $R$) + 4 * $n_{rollouts}$ (In each rollout there are 4 LLM Calls) + 2 (For final Syntax Correction and Deduplication). 

In our case, $n_{rollouts} = 4$, resulting in a maximum of 20 API calls to DeepSeek-R1 per signal. Since the generation of assertions for some signals might stop early due to syntax correctness, the actual number of API calls made is slightly less.

\begin{table}[h!]
\centering
\begin{tabular}{|l|c|c|c|c|}
\hline
\textbf{Design} & \textbf{\# Signals} & \textbf{Max} & \textbf{Total} & \textbf{Cost (USD)} \\
&&\textbf{API Calls}&\textbf{LLM Calls}&\\
\hline
\textbf{I2C}       & 23 & 460 & 450 & \$11.16 \\
\textbf{RV-Timer}  & 10 & 200  & 131 & \$4.85 \\
\hline
\textbf{Total} & \textbf{33} & 660 & \textbf{581} & \textbf{\$16.01} \\
\hline
\end{tabular}
\caption{Cost analysis of DeepSeek-R1 API usage}
\label{tab:cost_analysis}
\end{table}

\end{document}